\documentclass[
]{ceurart}

\usepackage{listings}
\begin{document}

\copyrightyear{2022}
\copyrightclause{Copyright for this paper by its authors.
  Use permitted under Creative Commons License Attribution 4.0
  International (CC BY 4.0).}

\conference{CLEF 2022: Conference and Labs of the Evaluation Forum, 
    September 5--8, 2022, Bologna, Italy}


\title{A Late Fusion Framework with Multiple Optimization Methods for Media Interestingness}

\author[1]{Maria Shoukat}[%
email=marvi1708@hotmail.com 
]
\address[1]{Department of Computer Systems Engineering, University of Engineering and Technology, Peshawar, Pakistan}

\author[1]{Khubaib Ahmad}[%
email=imkhubaib1999@gmail.com
]
\address[2]{Department of Computer Science, Munster Technological University, Cork, Ireland}
\author[1]{Naina Said}[%
email=nainasaid@uetpeshawar.edu.pk 
]
\author[1]{Nasir Ahmad}[%
email=n.ahmad@uetpeshawar.edu.pk 
]
\author[2]{Mohammed Hasanuzzaman}[%
email=mohammed.hasanuzzaman@mtu.ie
]
\author[2]{Kashif Ahmad}[%
email=kashif.ahmad@mtu.ie
]

\begin{abstract}
 The recent advancement in Multimedia Analytical, Computer Vision (CV), and Artificial Intelligence (AI) algorithms resulted in several interesting tools allowing an automatic analysis and retrieval of multimedia content of users' interests. However, retrieving the content of interest generally involves analysis and extraction of semantic features, such as emotions and interestingness-level. The extraction of such meaningful information is a complex task and generally, the performance of individual algorithms is very low. One way to enhance the performance of the individual algorithms is to combine the predictive capabilities of multiple algorithms using fusion schemes. This allows the individual algorithms to complement each other, leading to improved performance. This paper proposes several fusion methods for the media interestingness score prediction task introduced in CLEF Fusion 2022. The proposed methods include both a naive fusion scheme, where all the inducers are treated equally and a merit-based fusion scheme where multiple weight optimization methods are employed to assign weights to the individual inducers. In total, we used six optimization methods including a Particle Swarm Optimization (PSO), a Genetic Algorithm (GA), Nelder-Mead, Trust Region Constrained (TRC), and Limited-memory Broyden–Fletcher–Goldfarb–Shanno Algorithm (LBFGSA), and Truncated Newton Algorithm (TNA). Overall better results are obtained with PSO and TNA achieving 0.109 mean average precision @10. The task is complex and generally, scores are low. We believe the presented analysis will provide a baseline for future research in the domain.
\end{abstract}


\maketitle

\section{Introduction}
In the modern world, thanks to social media and other multimedia content sharing platforms, we have access to a huge amount of multimedia content. However, accessing multimedia content of interest generally involves processing, analyzing, and filtering a huge amount of data. Filtering and retrieval of multimedia content of interest require specialized tools to analyze and extract semantic features/meanings from the content \cite{demarty2017predicting}. 

Thanks to the recent development in Multimedia Analytics, Computer Vision (CV) and Artificial Intelligence (AI) techniques, different semantic notions, such as sentiments \cite{hassan2022visual}, emotions \cite{bhattacharya2021exploring}, and interestingness-level \cite{kiziltepe2021annotated}, can be extracted from multimedia content. More recently, Deep Neural Networks (DNNs) have shown tremendous predictive capabilities in various multimedia content analysis tasks. Despite the proven performances, there are several tasks where a single Neural Network is not enough to accurately extract meaningful insights precisely. In order to increase the performance of individual models, researchers have been exploring the so called "fusion" techniques allowing multiple models to complement each others in such complex tasks \cite{ahmad2018ensemble}. The fusion techniques allow to combine multiple models to achieve higher accuracy compared to the individual models. Two popular methods to combine these models are the early fusion and late fusion techniques. In early fusion, the separate raw data is integrated into a unified representation before the learning process. On the other hand, in case of late fusion, fusion is performed at the decision level i.e, the output of different predictors is combined after the learning process to create a new and improved super predictor. The literature has reported the effectiveness of fusion technique in several applications, such as natural disasters analysis \cite{said2018deep}, event recognition \cite{ahmad2018ensemble}, data analytics \cite{wang2021survey}.

This paper is based on one of the tasks in ImageCLEF 2022 \cite{ImageCLEF2022}, which is a benchmark competition for image retrieval tasks. This year ImageCLEF proposes four different tasks. However, the work is based on Image Interestingness CLEFfusion 2022 task \cite{ImageCLEF2022Fusion}. This is a regression task that aims at the prediction of image interestingness score.  The task mainly focuses on the fusion of different inducers, whose scores are already provided, to jointly predict the interestingness of visual content. In this work, we propose both a naive fusion, where all inducers are assigned equal weights, and merit-based fusion techniques with optimized weights to combine the scores of the individual inducers for better prediction. For the merit-based fusion, we employ five different techniques to assign weights to the 29 inducers provided in the dataset by the organizers. These methods include evolutionary algorithms, namely Particle Swarm Optimization (PSO) and a Genetic algorithm (GA) based methods, Trust Region Constrained Optimization,  Limited-memory Broyden Fletcher Goldfarb Shann (LMBFGS) method, and Truncated Newton Algorithm (TNA) method.

The rest of the paper is organized as follows : Section \ref{sec:related_work} presents the overview of the related literature. Section \ref{sec:fusion_techniques} gives details about different fusion techniques that have been used in this research work. Section \ref{sec:results} provides the experimental results. Finally, Section \ref{sec:conclusion} concludes the work and outlines the future directions.

\section{Related Work}
\label{sec:related_work}
Media interestingness prediction, which involves an automatic analysis of multimedia content for the identification of relevant content of users' interest, got great attention from the community over the last few years \cite{constantin2021visual}. It plays a vital role in several applications, such as image retrieval and recommendation and media summarization, etc. In the literature, the topic has been analyzed from two different perspectives including psychological and computational aspects of media interestingness \cite{constantin2021visual}. The first part mainly focuses on psychological studies involving theoretical analysis and reports on human emotions, choices, and interests. For instance, Silvia et al. \cite{silvia2009looking} linked interestingness level with emotions by providing a detailed analysis and overview of some unusual aesthetic emotions. The computational methods on the other hand involve multimedia analytics, CV, and ML techniques to analyze and extract semantic features from multimedia content for the prediction of interestingness level \cite{constantin2022exploring}. Extensive research exploring different aspects of the topic has been carried out in this direction. For instance, Liu et al. \cite{liu2017multi} analyzed the importance of feature extraction for the task by proposing a multi-view manifold learning framework. To this aim, the authors mapped multi-view data to a single common space by considering cross-view correlation to preserve the geometric structure and interestingness information. Wang et al. \cite{wang2018video} discussed other two important aspects of media interestingness namely comparison information, and evaluation metric optimization. Despite sufficient improvement, the performance of most of the algorithms is not good as compared to other computer vision tasks. 

As part of the efforts to improve the performances of media interestingness frameworks, a vast majority of the recent works rely on multiple models. To this aim, different fusion techniques have been incorporated to jointly employ multiple models for the task. For instance, Constantin et al. \cite{constantin2022exploring} proposed a deep fusion ensemble framework by exploring the potential of several deep networks including dense, attention, convolutional, and cross-space-fusion networks. Similarly, Almedia et al. \cite{almeida2017rank} proposed a late fusion framework employing multiple ranking models trained on multimodal features for media interestingness score prediction. 

In this work, we explore the potential of merit-based fusion by combining the predictions of several inducers using different weight optimization methods. 

\section{Methodology}
\begin{figure}[!h]
\centering
\includegraphics[width=.8\linewidth]{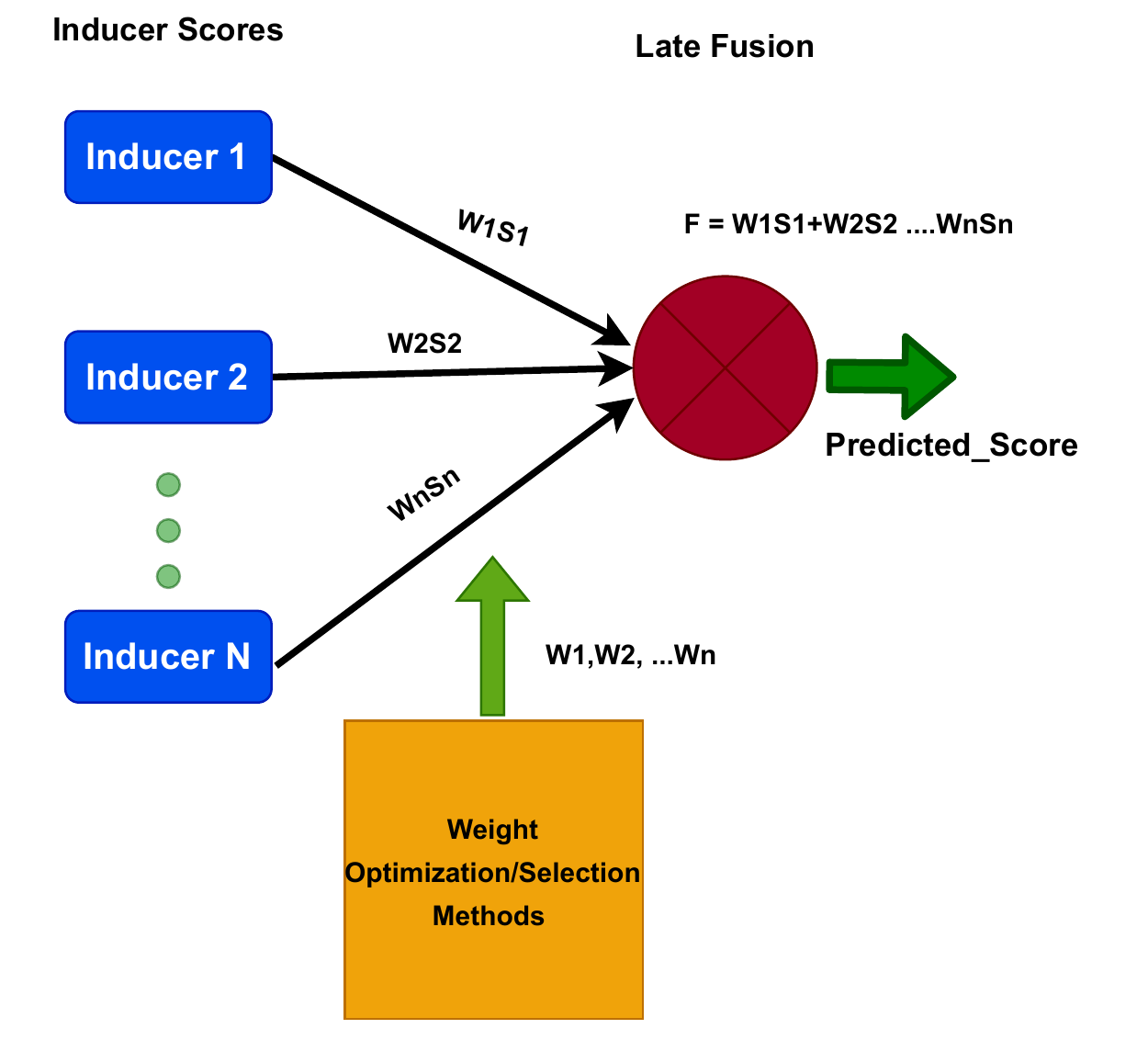}
\caption{Block diagram of the proposed methodology.}
	\label{fig:methodology}
\end{figure}

Figure \ref{fig:methodology} provides the block diagram of the methodology adopted in this work. There are two main components of the methodology, namely (i) prediction by individual inducers, and (ii) fusion of the score obtained with the individual inducers for joint prediction. Our contribution mainly lies in the fusion part, where we employed several weight selection/optimization techniques to assign weights to the inducers. The scores of the individual inducers are already provided by the task organizers.  In the next subsections, we provide a detailed description of all the methods.

\subsection{Inducer Scores and Pre-processing}
The inducers' scores are provided by the task organizers. In total, the task organizers provided prediction scores of 29 inducers. For the majority of the inducers, the prediction scores were floating numbers between 0 and 1. However, the interestingness scores for some of the inducers were out of range. To combine the scores of the inducers properly, all the values should be in the same range. To this aim, before combing the scores in a late fusion, we normalized the inducers' scores to bring them to the same range.  

\subsection{Fusion Techniques}
\label{sec:fusion_techniques}
In this work, we mainly focused on late fusion techniques where we tried several weight optimization methods to obtain a combination of weights that provides highest interestingness score. To this aim, we picked several methods with proven performances in similar applications \cite{ahmad2022merit,ahmad2018ensemble}. As a baseline, we also considered a naive fusion method where equal weights are assigned to the inducers. Our late fusion method is represented by Equation \ref{sec:fusion_techniques}.

\begin{equation}
\label{eqn:fusion}
S_{c}=W_{1}S_{1}+W_{2}S_{2}+W_{3}S_{3}+....+W_{n}S_{n}
\end{equation}

In the equation, $S_{c}$ represents the combined interestingness score of different inducers, $S_{n}$ is the score of the nth model and $W_{n}$ is the corresponding weight used during the fusion. In our case, n=29. 
For the baseline, the weight $W$ for all the inducers is the same i.e., $W_{1} = W_{2} = W_{3} .... = W_{n}$. In case of merit based fusion, optimal weights are assigned to each individual inducer. In the next section, we provide the details of these optimization methods.

\subsubsection{Genetic Algorithm}
Our choice of Genetic Algorithm (GA) is based on our previous experience in similar applications \cite{ahmad2022merit,ahmad2020intelligent,ahmad2018ensemble}. GA, which is a meta-heuristic algorithm, is inspired by the natural evaluation process. In the natural evolution process, the fittest individuals of the current generation are firstly identified and then used for re-production in the next generation. A similar approach is adopted in GA-based optimization, where the algorithm searches for optimal value/set of values minimizing a given function also called fitness function. To this aim, GA requires a training procedure to determine optimal values. The process starts with a randomly selected generation of the population (i.e., a set of values). The fitness of every individual in the current population is evaluated using the fitness function after which individuals are selected for the next generation with a modified genome. The process is repeated until either the maximum number of generations is reached or a respectable fitness value is achieved. 

In this work, since we are dealing with a regression problem, the fitness function is based on Mean Squared Error (MSE) as shown in equation \ref{eqn:fitness}. Moreover, each possible combination (i.e., set of weights assigned to the 29 inducers) is a potential solution. The goal is to find the set of weights with minimum MSE.

\begin{equation}
\label{eqn:fitness}
 MSE = {\frac{1}{n}\Sigma_{i=1}^{n}(y_p -y_a)}^2
\end{equation}

In the above equation, $y_p$ represents the predicted interestingness score while $y_a$ is the ground truth. 
For the implementation we used a python open source library, namely geneticalgorithm\footnote{https://pypi.org/project/geneticalgorithm/}. As we have a total of 29 inducers so we kept the dimensions to 29. Moreover, we used 'real' for the variable type and the variable boundary fixed between 0 and 1.

\subsubsection{Particle Swarm Optimzation}
The second weight optimization/selection method employed in this work is based on PSO \cite{kennedy1995particle}. The optimization method is inspired by the flocking of birds. Unlike GA, PSO does not use mutation and crossover operations rather aims at an improvement to a candidate solution according to a pre-defined criterion, iteratively. There are three main steps involved in the process. These include an (i) evaluation of each candidate solution on the basis of fitness criteria, (ii) updates in personal best  and global best values and finally (iii) updating the position and velocity of each particle. 

In our case, each combination of the weights (i.e, 29 values to be assigned to the inducers) is a candidate solution. Moreover, the fitness function is based on MSE as shown in \ref{eqn:fitness}. For the implementation of the method, we used open source library namely pyswarm\footnote{https://pyswarms.readthedocs.io/en/latest/}. As this algorithm support bounds like GA, we set the lower bound to 0 and the upper bound to be 1. Moreover, the maximum iterations hyperparameter is set to 10000 and the swarm size is kept at 300.


\subsubsection{Nelder Mead Algorithm}
Nelder Mead algorithm is a heuristic optimization technique and is appropriate for optimization problems where the gradient of the function is either unknown or cannot be reasonably computed. The algorithm can be used for both one dimensional and multi-dimensional optimization problems \cite{singer2009nelder}. The algorithm starts with randomly generated simplex with number of vertices= $n+1$ points for an n-dimensional optimization problem. At every iteration, the algorithm moves the simplex one vertex at a time towards an optimal region in the search space with a goal to minimize/maximize a certain objective function. At the end, the vertex of the simplex that yields that most optimal values is returned. For our experiments, $n=29$ and the objective function definition is the same as in \ref{eqn:fitness}. For the implementation of the method, we used a Python open source library, namely, SciPy\footnote{https://scipy.org/}. We set the value of absolute error in xopt between iterations that is acceptable for convergence to 1e-8 and the maximum iterations to 10000.


\subsubsection{Trust Region Constrained Optimization Algorithm }
Trust Region Constrained method belongs to the family of optimization methods that are based on trust regions. Trust regions-based methods solve optimization problems by defining a region around their current best solutions, where they can approximate the fitness function up to a certain extent. The methods then take a step forward within the region. In contrast to line-based solutions, the step size is determined beforehand of the improvement in the direction. At this stage, the model is considered to be a good representation of the original objective function if a significant decrease in the objective function is observed.

In this work, for the implementation of the method we used scipy library\footnote{https://scipy.org/}. The Trust Region Constrained algorithm requires initial weight values so for all the 29 inducers same value of 0.0345 is used. For the bounds, we used a lower bound of 0 and upper bound of 1. Moreover, we set the maximum iterations to 10000.


\subsubsection{Limited-memory Broyden–Fletcher–Goldfarb–Shanno Algorithm}
The Broyden, Fletcher, Goldfarb, and Shanno, or BFGS Algorithm, is a local search optimization algorithm. It falls under the  category of a Quasi-Newton optimization methods which deal with the optimization of second order derivative of the objective function. These type of algorithms are suitable for optimization problems where the second order derivatives can not be reasonably quantified. Unlike the first order methods which make use of the first order derivative to find the optimal values of the objective function, these algorithm rely on second order derivatives. For a multivariate function, the second derivatives of all the input variables are maintained in a matrix called Hessian. In order to find the optimal values of the objective function, the BFGS algorithm calculates the inverse of this matrix. This is done by approximating the inverse using gradient thereby eliminating the need for inverse calculation at each step of the algorithm. The size of the Hessian and its inverse is proportional to the number of input parameters to the objective function. For a function with many input parameters, the BFGS then becomes impractical due to very high memory demands. Therefore, a variant of BFGS called Limited BFGS is utilized in this work. This method does not require storing the entire approximation of the inverse matrix. 

The definition of the objective function for our experiments is the same as given in \ref{eqn:fitness}. For implementation of the method, we used a Python open source library, namely, SciPy\footnote{https://scipy.org/}. Similar to Trust Region Constrained Optimization algorithm, the Limited-memory Broyden–Fletcher–Goldfarb–Shanno algorithm also requires initial weight values that are set at 0.0345 for all the 29 inducers. Moreover, the value of absolute error is set to 1e-8 and the maximum iterations to 10000. For the bounds, we set the lower and upper bounds to 0 and 1, respectively.


\subsubsection{Truncated Newton Algorithm}
The method is also called Hessian-free optimization algorithm and is more suitable for applications involving a large numbers of independent variables \cite{martens2010deep}. The method uses an iterative process to solve the Newton's equation, which involves finding the roots of a differentiable function, for updating the parameters of the cost function. The term ''truncated'' refers to the fact that the inner solver is run for a limited number of iterations, which means that the algorithm needs to produce good approximation in limited iterations. 
The definition of the objective function for our experiments is the same as given in
2. For implementation of the method, we used a Python open source library, namely,
SciPy. The Truncated Newton algorithm also requires initial weight values, which are set at 0.0345 for all the 29 inducers. Moreover, we set the value of absolute error in xopt between
iterations that are acceptable for convergence to 1e-8 and the maximum iterations to 10000. The lower and upper bounds are set at 0 and 1, respectively.


\section{Experiments and Results}
\label{sec:results}

\subsection{Dataset}
The individual inducers' scores, which are provided by the task organizers, are extracted from the Interestingness10k dataset \cite{constantin2021visual}. In total, prediction scores for 2435 images are provided from 29 inducers, each representing the visual interestingness-level for the images. The dataset is provided in two seperate sets namely (i) development set, and (ii) test set. The development set is composed of the scores from all of the inducers for 1877 images while the test set covers 558 images only. In the development set, each data sample for each inducer provides four values including a video ID, image ID, classification score (0 or 1), and predicted interestingness score by the inducer. 

\subsection{Experimental Results}
Table \ref{tab:results} provides the official results of the proposed methods in terms of mean average precision at the cutoff 10 (MAP@10). One of the main objectives of the experiments is to evaluate the potential of the different state-of-the-art optimization methods in this application. 

As expected, overall lower results are obtained with the baseline method where all the inducers are treated equally by assigning them equal weights. Though there is no significant, difference in scores obtained between the least performing and the baseline, the merit-based fusion scheme seems more promising compared to simply averaging the individual scores. As far as the performances of the merit-based fusion methods are concerned, overall better results are obtained with PSO and TNC methods. One of the key advantages of TNC is its optimization capabilities in dealing with functions involving independent variables. This could be one of the main reasons for its better performance in the application as all the inducers are treated independently in the task. The highest score obtained in this work is 0.109 MAP@10, which indicates the complexity of the task.

\begin{table}[]
\centering
\label{tab:results}
\caption{Experimental results in terms of mean average precision at the cutoff 10 MAP@10.}
\begin{tabular}{|c|c|}
\hline
\textbf{Fusion Method} & \textbf{MAP@10} \\ \hline
 Equal Weights& 0.081 \\ \hline
Trust-Constr weighted Fusion & 0.095 \\ \hline
PSO weighted Fusion& \textbf{ 0.109}\\ \hline
GA weighted Fusion&  0.093\\ \hline
LBFGSB weighted Fusion&  0.095\\ \hline
Nelder Mead weighted Fusion&  0.090\\ \hline
TNC weighted Fusion& \textbf{ 0.109}\\ \hline
\end{tabular}
\end{table}

\section{Conclusions and Future Work}
\label{sec:conclusion}
In this paper, we presented the experimental results of multiple fusion techniques for the media interestingness task presented in CLEF Fusion 2022. We used both a naive fusion scheme and merit-based fusion methods. Overll the results are much lower on the task compared to other computer vision tasks, which shows the complexity of the task. During the experiments, we observed better results for a merit-based fusion scheme where different weight optimization techniques are employed to assign weights to the individual inducers based on their performances. This verifies our assumption that individual performance should be considered in combining the prediction scores of the individual inducers.

In the future, we want to further explore different aspects of the application to further enhance the results. One potential direction could be an intelligent selection among the inducers instead of considering all of them. 


\bibliography{references}

\begin{thebibliography}{20}
\expandafter\ifx\csname natexlab\endcsname\relax\def\natexlab#1{#1}\fi
\providecommand{\url}[1]{\texttt{#1}}
\providecommand{\href}[2]{#2}
\providecommand{\path}[1]{#1}
\providecommand{\DOIprefix}{doi:}
\providecommand{\ArXivprefix}{arXiv:}
\providecommand{\URLprefix}{URL: }
\providecommand{\Pubmedprefix}{pmid:}
\providecommand{\doi}[1]{\href{http://dx.doi.org/#1}{\path{#1}}}
\providecommand{\Pubmed}[1]{\href{pmid:#1}{\path{#1}}}
\providecommand{\bibinfo}[2]{#2}
\ifx\xfnm\relax \def\xfnm[#1]{\unskip,\space#1}\fi
\bibitem[{Demarty et~al.(2017)Demarty, Sj{\"o}berg, Constantin, Duong, Ionescu,
  Do, and Wang}]{demarty2017predicting}
\bibinfo{author}{C.-H. Demarty}, \bibinfo{author}{M.~Sj{\"o}berg},
  \bibinfo{author}{M.~G. Constantin}, \bibinfo{author}{N.~Q. Duong},
  \bibinfo{author}{B.~Ionescu}, \bibinfo{author}{T.-T. Do},
  \bibinfo{author}{H.~Wang},
\newblock \bibinfo{title}{Predicting interestingness of visual content},
\newblock in: \bibinfo{booktitle}{Visual content indexing and retrieval with
  psycho-visual models}, \bibinfo{publisher}{Springer}, \bibinfo{year}{2017},
  pp. \bibinfo{pages}{233--265}.
\bibitem[{Hassan et~al.(2022)Hassan, Ahmad, Hicks, Halvorsen, Al-Fuqaha, Conci,
  and Riegler}]{hassan2022visual}
\bibinfo{author}{S.~Z. Hassan}, \bibinfo{author}{K.~Ahmad},
  \bibinfo{author}{S.~Hicks}, \bibinfo{author}{P.~Halvorsen},
  \bibinfo{author}{A.~Al-Fuqaha}, \bibinfo{author}{N.~Conci},
  \bibinfo{author}{M.~Riegler},
\newblock \bibinfo{title}{Visual sentiment analysis from disaster images in
  social media},
\newblock \bibinfo{journal}{Sensors} \bibinfo{volume}{22}
  (\bibinfo{year}{2022}) \bibinfo{pages}{3628}.
\bibitem[{Bhattacharya et~al.(2021)Bhattacharya, Gupta, and
  Yang}]{bhattacharya2021exploring}
\bibinfo{author}{P.~Bhattacharya}, \bibinfo{author}{R.~K. Gupta},
  \bibinfo{author}{Y.~Yang},
\newblock \bibinfo{title}{Exploring the contextual factors affecting multimodal
  emotion recognition in videos},
\newblock \bibinfo{journal}{IEEE Transactions on Affective Computing}
  (\bibinfo{year}{2021}).
\bibitem[{Kiziltepe et~al.(2021)Kiziltepe, Sweeney, Constantin, Doctor,
  de~Herrera, Demarty, Healy, Ionescu, and Smeaton}]{kiziltepe2021annotated}
\bibinfo{author}{R.~S. Kiziltepe}, \bibinfo{author}{L.~Sweeney},
  \bibinfo{author}{M.~G. Constantin}, \bibinfo{author}{F.~Doctor},
  \bibinfo{author}{A.~G.~S. de~Herrera}, \bibinfo{author}{C.-H. Demarty},
  \bibinfo{author}{G.~Healy}, \bibinfo{author}{B.~Ionescu},
  \bibinfo{author}{A.~F. Smeaton},
\newblock \bibinfo{title}{An annotated video dataset for computing video
  memorability},
\newblock \bibinfo{journal}{Data in Brief} \bibinfo{volume}{39}
  (\bibinfo{year}{2021}) \bibinfo{pages}{107671}.
\bibitem[{Ahmad et~al.(2018)Ahmad, Mekhalfi, Conci, Melgani, and
  Natale}]{ahmad2018ensemble}
\bibinfo{author}{K.~Ahmad}, \bibinfo{author}{M.~L. Mekhalfi},
  \bibinfo{author}{N.~Conci}, \bibinfo{author}{F.~Melgani},
  \bibinfo{author}{F.~D. Natale},
\newblock \bibinfo{title}{Ensemble of deep models for event recognition},
\newblock \bibinfo{journal}{ACM Transactions on Multimedia Computing,
  Communications, and Applications (TOMM)} \bibinfo{volume}{14}
  (\bibinfo{year}{2018}) \bibinfo{pages}{1--20}.
\bibitem[{Said et~al.(2018)Said, Pogorelov, Ahmad, Riegler, Ahmad, Ostroukhova,
  Halvorsen, and Conci}]{said2018deep}
\bibinfo{author}{N.~Said}, \bibinfo{author}{K.~Pogorelov},
  \bibinfo{author}{K.~Ahmad}, \bibinfo{author}{M.~Riegler},
  \bibinfo{author}{N.~Ahmad}, \bibinfo{author}{O.~Ostroukhova},
  \bibinfo{author}{P.~Halvorsen}, \bibinfo{author}{N.~Conci},
\newblock \bibinfo{title}{Deep learning approaches for flood classification and
  flood aftermath detection.},
\newblock in: \bibinfo{booktitle}{MediaEval}, \bibinfo{year}{2018}.
\bibitem[{Wang(2021)}]{wang2021survey}
\bibinfo{author}{Y.~Wang},
\newblock \bibinfo{title}{Survey on deep multi-modal data analytics:
  Collaboration, rivalry, and fusion},
\newblock \bibinfo{journal}{ACM Transactions on Multimedia Computing,
  Communications, and Applications (TOMM)} \bibinfo{volume}{17}
  (\bibinfo{year}{2021}) \bibinfo{pages}{1--25}.
\bibitem[{Ionescu et~al.(2022)Ionescu, M\"uller, Peteri, Ruckert, {Ben Abacha},
  de~Herrera, Friedrich, Bloch, Br\"ungel, Idrissi-Yaghir, Sch\"afer,
  Kozlovski, Cid, Kovalev, Stefan, Constantin, Dogariu, Popescu,
  Deshayes-Chossart, Schindler, Chamberlain, Campello, and
  Clark}]{ImageCLEF2022}
\bibinfo{author}{B.~Ionescu}, \bibinfo{author}{H.~M\"uller},
  \bibinfo{author}{R.~Peteri}, \bibinfo{author}{J.~Ruckert},
  \bibinfo{author}{A.~{Ben Abacha}}, \bibinfo{author}{A.~G.~S. de~Herrera},
  \bibinfo{author}{C.~M. Friedrich}, \bibinfo{author}{L.~Bloch},
  \bibinfo{author}{R.~Br\"ungel}, \bibinfo{author}{A.~Idrissi-Yaghir},
  \bibinfo{author}{H.~Sch\"afer}, \bibinfo{author}{S.~Kozlovski},
  \bibinfo{author}{Y.~D. Cid}, \bibinfo{author}{V.~Kovalev},
  \bibinfo{author}{L.-D. Stefan}, \bibinfo{author}{M.~G. Constantin},
  \bibinfo{author}{M.~Dogariu}, \bibinfo{author}{A.~Popescu},
  \bibinfo{author}{J.~Deshayes-Chossart}, \bibinfo{author}{H.~Schindler},
  \bibinfo{author}{J.~Chamberlain}, \bibinfo{author}{A.~Campello},
  \bibinfo{author}{A.~Clark},
\newblock \bibinfo{title}{{Overview of the ImageCLEF 2022}: {Multimedia
  Retrieval in Medical, Social Media and Nature Applications}},
\newblock in: \bibinfo{booktitle}{Experimental IR Meets Multilinguality,
  Multimodality, and Interaction}, Proceedings of the 13th International
  Conference of the CLEF Association (CLEF 2022), \bibinfo{publisher}{{LNCS}
  Lecture Notes in Computer Science, Springer}, \bibinfo{address}{Bologna,
  Italy}, \bibinfo{year}{2022}.
\bibitem[{Stefan et~al.(2022)Stefan, Constantin, Dogariu, and
  Ionescu}]{ImageCLEF2022Fusion}
\bibinfo{author}{L.-D. Stefan}, \bibinfo{author}{M.~G. Constantin},
  \bibinfo{author}{M.~Dogariu}, \bibinfo{author}{B.~Ionescu},
\newblock \bibinfo{title}{Overview of imagecleffusion 2022 task - ensembling
  methods for media interestingness prediction and result diversification},
\newblock in: \bibinfo{booktitle}{CLEF2022 Working Notes}, {CEUR} Workshop
  Proceedings, \bibinfo{publisher}{CEUR-WS.org}, \bibinfo{address}{Bologna,
  Italy}, \bibinfo{year}{2022}.
\bibitem[{Constantin et~al.(2021)Constantin, {\c{S}}tefan, Ionescu, Duong,
  Demarty, and Sj{\"o}berg}]{constantin2021visual}
\bibinfo{author}{M.~G. Constantin}, \bibinfo{author}{L.-D. {\c{S}}tefan},
  \bibinfo{author}{B.~Ionescu}, \bibinfo{author}{N.~Q. Duong},
  \bibinfo{author}{C.-H. Demarty}, \bibinfo{author}{M.~Sj{\"o}berg},
\newblock \bibinfo{title}{Visual interestingness prediction: A benchmark
  framework and literature review},
\newblock \bibinfo{journal}{International Journal of Computer Vision}
  \bibinfo{volume}{129} (\bibinfo{year}{2021}) \bibinfo{pages}{1526--1550}.
\bibitem[{Silvia(2009)}]{silvia2009looking}
\bibinfo{author}{P.~J. Silvia},
\newblock \bibinfo{title}{Looking past pleasure: anger, confusion, disgust,
  pride, surprise, and other unusual aesthetic emotions.},
\newblock \bibinfo{journal}{Psychology of Aesthetics, Creativity, and the Arts}
  \bibinfo{volume}{3} (\bibinfo{year}{2009}) \bibinfo{pages}{48}.
\bibitem[{Constantin et~al.(2022)Constantin, {\c{S}}tefan, and
  Ionescu}]{constantin2022exploring}
\bibinfo{author}{M.~G. Constantin}, \bibinfo{author}{L.-D. {\c{S}}tefan},
  \bibinfo{author}{B.~Ionescu},
\newblock \bibinfo{title}{Exploring deep fusion ensembling for automatic visual
  interestingness prediction},
\newblock in: \bibinfo{booktitle}{Human Perception of Visual Information},
  \bibinfo{publisher}{Springer}, \bibinfo{year}{2022}, pp.
  \bibinfo{pages}{33--58}.
\bibitem[{Liu et~al.(2017)Liu, Gu, Cheung, and Hua}]{liu2017multi}
\bibinfo{author}{Y.~Liu}, \bibinfo{author}{Z.~Gu}, \bibinfo{author}{Y.-m.
  Cheung}, \bibinfo{author}{K.~A. Hua},
\newblock \bibinfo{title}{Multi-view manifold learning for media
  interestingness prediction},
\newblock in: \bibinfo{booktitle}{Proceedings of the 2017 ACM on International
  Conference on Multimedia Retrieval}, \bibinfo{year}{2017}, pp.
  \bibinfo{pages}{308--314}.
\bibitem[{Wang et~al.(2018)Wang, Chen, Zhao, and Jin}]{wang2018video}
\bibinfo{author}{S.~Wang}, \bibinfo{author}{S.~Chen},
  \bibinfo{author}{J.~Zhao}, \bibinfo{author}{Q.~Jin},
\newblock \bibinfo{title}{Video interestingness prediction based on ranking
  model},
\newblock in: \bibinfo{booktitle}{Proceedings of the joint workshop of the 4th
  workshop on affective social multimedia computing and first multi-modal
  affective computing of large-scale multimedia data}, \bibinfo{year}{2018},
  pp. \bibinfo{pages}{55--61}.
\bibitem[{Almeida et~al.(2017)Almeida, Valem, and Pedronette}]{almeida2017rank}
\bibinfo{author}{J.~Almeida}, \bibinfo{author}{L.~P. Valem},
  \bibinfo{author}{D.~C. Pedronette},
\newblock \bibinfo{title}{A rank aggregation framework for video
  interestingness prediction},
\newblock in: \bibinfo{booktitle}{International conference on image analysis
  and processing}, \bibinfo{organization}{Springer}, \bibinfo{year}{2017}, pp.
  \bibinfo{pages}{3--14}.
\bibitem[{Ahmad et~al.(2022)Ahmad, Ayub, Ahmad, Khan, Ahmad, and
  Al-Fuqaha}]{ahmad2022merit}
\bibinfo{author}{K.~Ahmad}, \bibinfo{author}{M.~A. Ayub},
  \bibinfo{author}{K.~Ahmad}, \bibinfo{author}{J.~Khan},
  \bibinfo{author}{N.~Ahmad}, \bibinfo{author}{A.~Al-Fuqaha},
\newblock \bibinfo{title}{Merit-based fusion of nlp techniques for instant
  feedback on water quality from twitter text},
\newblock \bibinfo{journal}{arXiv preprint arXiv:2202.04462}
  (\bibinfo{year}{2022}).
\bibitem[{Ahmad et~al.(2020)Ahmad, Khan, and Al-Fuqaha}]{ahmad2020intelligent}
\bibinfo{author}{K.~Ahmad}, \bibinfo{author}{K.~Khan},
  \bibinfo{author}{A.~Al-Fuqaha},
\newblock \bibinfo{title}{Intelligent fusion of deep features for improved
  waste classification},
\newblock \bibinfo{journal}{IEEE access} \bibinfo{volume}{8}
  (\bibinfo{year}{2020}) \bibinfo{pages}{96495--96504}.
\bibitem[{Kennedy and Eberhart(1995)}]{kennedy1995particle}
\bibinfo{author}{J.~Kennedy}, \bibinfo{author}{R.~Eberhart},
\newblock \bibinfo{title}{Particle swarm optimization},
\newblock in: \bibinfo{booktitle}{Proceedings of ICNN'95-international
  conference on neural networks}, volume~\bibinfo{volume}{4},
  \bibinfo{organization}{IEEE}, \bibinfo{year}{1995}, pp.
  \bibinfo{pages}{1942--1948}.
\bibitem[{Singer and Nelder(2009)}]{singer2009nelder}
\bibinfo{author}{S.~Singer}, \bibinfo{author}{J.~Nelder},
\newblock \bibinfo{title}{Nelder-mead algorithm},
\newblock \bibinfo{journal}{Scholarpedia} \bibinfo{volume}{4}
  (\bibinfo{year}{2009}) \bibinfo{pages}{2928}.
\bibitem[{Martens et~al.(2010)}]{martens2010deep}
\bibinfo{author}{J.~Martens}, et~al.,
\newblock \bibinfo{title}{Deep learning via hessian-free optimization.},
\newblock in: \bibinfo{booktitle}{ICML}, volume~\bibinfo{volume}{27},
  \bibinfo{year}{2010}, pp. \bibinfo{pages}{735--742}.

\end{thebibliography}

\appendix



\end{document}